# Can Visual Dialogue Models Do Scorekeeping? Exploring How Dialogue Representations Incrementally Encode Shared Knowledge

**Brielen Madureira** **David Schlangen**
Computational Linguistics
Department of Linguistics
University of Potsdam, Germany
{madureiralasota, david.schlangen}@uni-potsdam.de

## Abstract

Cognitively plausible visual dialogue models should keep a *mental scoreboard* of shared established facts in the dialogue context. We propose a theory-based evaluation method for investigating to what degree models pretrained on the VisDial dataset incrementally build representations that appropriately do *scorekeeping*. Our conclusion is that the ability to make the distinction between shared and privately known statements along the dialogue is moderately present in the analysed models, but not always incrementally consistent, which may partially be due to the limited need for *grounding interactions* in the original task.

## 1 Introduction

"There's a cute dog outside!" you say on the phone to your friend. "Sweet. What colour is the dog?", they say. "What dog?" you reply – and your friend is rightfully confused. With your first utterance, you have committed yourself to there being a dog; a commitment you can't just simply ignore later on. Models of dialogue from linguistics and psycholinguistics take this process of *grounding* or *scorekeeping*—making propositions mutual knowledge—to be an elementary fact about dialogue (Lewis, 1979; Clark and Brennan, 1991).

In this short paper, we investigate whether recent NLP models of visual dialogue capture this process. Specifically, we use the VisDial dataset (Das et al., 2017a), which consists of dialogues in English about an image in an asymmetric setting similar to that from the first paragraph, and derive from it diagnostic propositions that should be considered mutual knowledge at a given point in the dialogue, and others whose truth value is only known to one participant at the given time. We then probe dialogue representations built by models pretrained on the VisDial task for whether they correctly track the participants' knowledge and commitments.

## 2 Related Literature

Representing dialogue context implicitly as the continuous hidden states of neural networks trained in an end-to-end fashion has been a prevailing practice since the works of Vinyals and Le (2015), Sordoni et al. (2015) and Serban et al. (2016). This paradigm also enables multimodal input like images to be easily integrated (Shekhar et al., 2019b). However, there is evidence that the human ability of *collaborative grounding* still lacks in such models, in part due to the limitations of training regimes and datasets (Benotti and Blackburn, 2021).

We witness extensive efforts to look into how these models encode and make use of dialogue history, capture salient information and produce visually grounded representations (Sankar et al., 2019; Agarwal et al., 2020; Greco et al., 2020a,b). The analysis and evaluation of current dialogue models (as Hupkes et al. (2018a), Shekhar et al. (2019a), Parthasarathi et al. (2020), Saleh et al. (2020), Wu and Xiong (2020), *inter alia*) often rely on diagnostic classifiers (Hupkes et al., 2018b) and probing tasks (Belinkov and Glass, 2019), common tools to examine whether representations built by neural networks encode linguistic information.

Another purposeful area of research on dialogue revolves around inference. Zhang and Chai (2009, 2010) discuss *conversation entailment*, *i.e.* determining whether a conversation discourse entails a hypothesis. Annotating or generating entailments, contradictions and neutral statements in dialogue datasets is usual in recent works (Welleck et al., 2019; Dziri et al., 2019; Galetzka et al., 2021).

With insights from these three pillars, we propose a probing task for *scorekeeping* (Lewis, 1979) on visual dialogues, formalised in the next section.

## 3 Problem Statement

Based on the premise that humans keep a *mental scoreboard* of presupposed propositions and per-missible courses of action as a function of what has been stated in a conversation (Lewis, 1979) and on the public/private dichotomy discussed in Ginzburg (2012), we propose a formalisation for the "kinematics of scorekeeping" (Lewis, 1979) on VisDial.

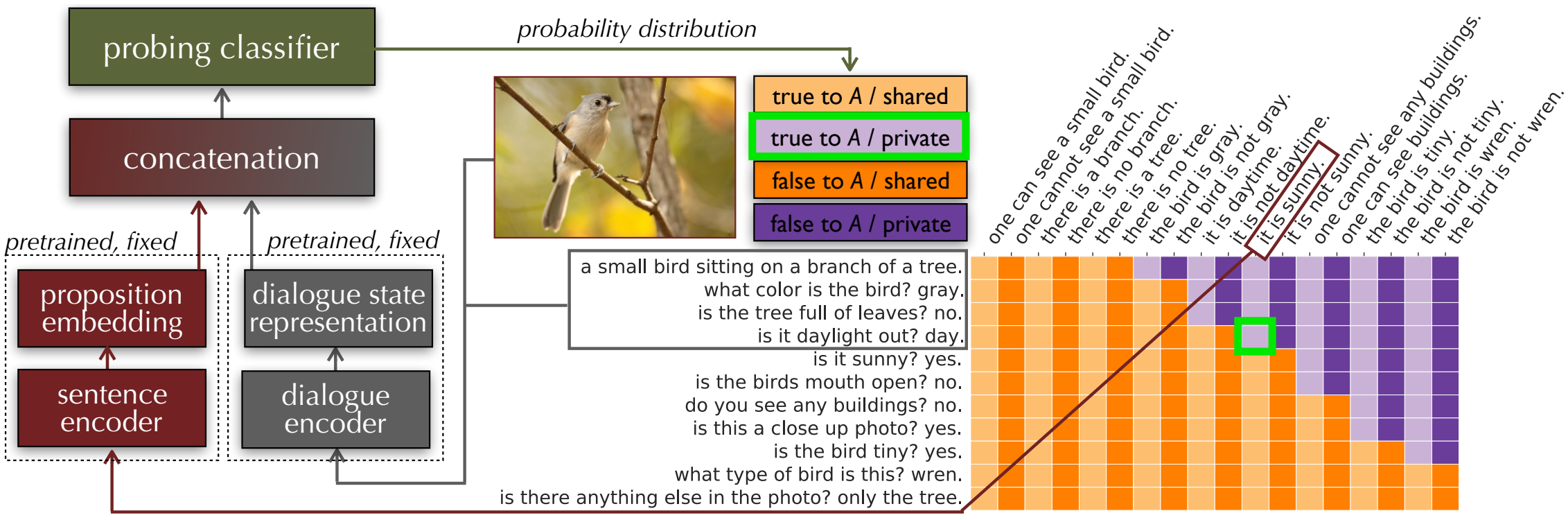

Figure 1: A scoreboard representation with generated propositions for a dialogue and architecture of the classifier. It represents the proposition *it is sunny* being correctly classified as (true to $A$, private) at turn 3. From VisDial training set, ID 8778 (CC-BY 4.0), photo 176904 from MS COCO dataset, ↪Tufted Titmouse by Matt Tillett (CC-BY 2.0).

Each dialogue in the VisDial dataset is a tuple $D = (I, Q, A, T, P)$ representing an interaction between a questioner $Q$ and an answerer $A$. They exchange turns $T$, which establish propositions $P$, about a scene depicted in an image $I$. $A$ sees $I$, but $Q$ does not. Both are provided with a caption $K$, which for simplicity we take to be the first turn of $A$, $t_0 = K$; other turns comprise a question and an answer, $t_i = (q_i, a_i)$, so that $T = (t_i)_{i=0}^{10}$ (as dialogues have 10 turns).[1]

We assume that: i) $A$ does not lie about their interpretation of the image; ii) $Q$ does not ask redundant questions; and iii) a fact disclosed by $A$ immediately becomes a shared commitment, even though in reality this is not always the case (*e.g.* when a misunderstanding happens). Under these assumptions, each $t_i$ discloses a new fact $p^i$ (and its implications) about $A$'s judgement of the image that was unknown to $Q$ until $t_{i-1}$. $P$ is then defined as a set of $N$ propositions $\{p_1^i, p_2^i, \cdots, p_N^i\}$. Each $p_j^i$ is either the direct entailment of $t_i$ (that is, the expressed proposition), which is established by $A$ to be *true*, or its negation, which is established by $A$ to be *false*. The truth value of $p_j^i$ is known to $A$ throughout the dialogue, but only *privately* so for all $k < i$. It becomes *shared* between $A$ and $Q$ at $k = i$ and remains so until the end of the dialogue.[2]

With this in place, $A$'s scoreboard of a dialogue can be represented by a matrix $S_D$ with dimensions $|T| \times |P|$. Each element $s_{m,n}$ is a tuple $c \in C = \{$(true to $A$, private), (true to $A$, shared), (false to $A$, private), (false to $A$, shared)$\}$ representing the 'score' of proposition $p_n$ at turn $t_m$ as a class, like the example in Figure 1. Hence, the negation of a fact that $A$ considers true but has not been mentioned yet is labelled as (false to $A$, private).[3] That way, the scoreboard at a given turn $t$ is given by the $t$-th row in $S$ and the whole matrix helps visualising how the scoreboard is incrementally updated throughout $D$.

**Probing Task and Model**. We design a classification task to examine whether the continuous representations of pretrained visual dialogue models incrementally encode information about the scoreboard represented by $S$. The probing classifier is a function $f : P_D \times R_{D,t} \to C$, where $P_D$ is the set of propositions in a dialogue $D$, $R$ is the space of hidden representations of a visual dialogue encoder and $C$ are the scoreboard classes. Based on the probing classifier architecture in Hewitt and Liang (2019), we approximate $f$ as a neural network which maps a dialogue representation $r$ concatenated to a continuous representation $z$ of a proposition to a vector $v$ with a probability distribution over classes, $v = softmax(W_2 \sigma(W_1[r; z]))$ (bias term omitted), as illustrated in Figure 1. The class is then predicted with the *argmax* function.

## 4 Data

**Visual Dialogues and Encoders**. We use the VisDial dataset v.1.0 (Das et al., 2017a) and the three $Q$ and $A$ encoders (RL_DIV, SL and ICCV_RL) from Das et al. (2017b) and Murahari et al. (2019). The first work implemented an end-to-end model to train $A$ and $Q$ using reinforcement learning. The latter is a follow-up study that adds an auxiliary objective function to encourage $Q$ to ask more diverse questions.[4] The VisDial training set contains images from the MS COCO dataset (Lin et al., 2014). Proposition embeddings $z$ are built with Sentence-Transformers (Reimers and Gurevych, 2019).

[1] Except on VisDial test set, where $T < 10$.

[2] Although the set of statements about an image can be infinitely large, we limit $P$ to a finite set here by only considering explicitly disclosed facts (and their negation).

[3] The scoreboard for $Q$ is analogous, except that it cannot differentiate the true/false dimension of private propositions.

**Generating Probes**. The sets $P_D$ are programmatically generated by manipulating QA pairs using rules that identify common lexical and syntactic patterns in VisDial, in a similar fashion as Demszky et al. (2018) and Ribeiro et al. (2019). Whenever the pattern of a QA pair matches a rule, a *direct entailment* and a *direct contradiction* are generated, as those shown in Figure 1.[5]

**Dataset Construction**. We retrieve the pretrained dialogue context representations $R_D = \{r_l | 0 \leq l \leq 10\}$, where $r_l$ is the hidden state of the encoder after it processed the dialogue up to turn $l$ in $T$ (and the image and next question for $A$). We then pair elements in $R_D$ with the embeddings of the generated propositions $p^i_j$ in $P_D$, forming tuples $\{(r_l, p^i_j) | 0 \leq l \leq 10, 1 \leq j \leq N\}$ which are mapped to the corresponding class $c \in C$. The *true to* $A$ or *false to* $A$ status of a proposition $p^i_j$ remains fixed for all turns in $D$, since it refers to a fact (according to $A$'s beliefs) about the image, while the *private* status holds for $(r_0, p^i_j), \ldots, (r_{i-1}, p^i_j)$ and shifts to *shared* for $(r_i, p^i_j), \ldots, (r_{10}, p^i_j)$. The probing dataset is thus composed of datapoints $(r, p, c)_D$ for all $D$, for all turns' representations $r \in R_D$, for all $p \in P_D$. Propositions generated from captions are downsampled because they outnumber the other turns, resulting in too many propositions that are always shared. In order to avoid bias with respect to the true/false dimension, we sample the training set of propositions enforcing that each type appears as true to $A$ exactly the same number of times as it does as false to $A$ in different dialogues. Table 1 presents a summary (see Appendix for details).

## 5 Experiments

We train and test the classifier varying three aspects: i) $A$ or $Q$, ii) main task with all classes in $C$

[4]Code and model checkpoints available under a BSD license at https://github.com/vmurahari3/visdial-diversity.

[5]The rule-based approach can only generate subsets of the theoretical $P_D$, but in enough number for the probing task. See Appendix for details and examples.

| | train | valid | test |
|---|---|---|---|
| dialogues | 95,369 | 1,979 | 6,880 |
| propositions | 344,988 | 23,060 | 44,954 |
| proposition types | 27,011 | 12,048 | 19,183 |
| datapoints | 3,794,868 | 253,660 | 312,102 |
| vocab size | 2,709 | 2,168 | 2,922 |
| avg. $\|P_D\|$ | 3.61 | 11.65 | 6.53 |
| true to $A$ and private | 26.12 | 22.94 | 21.42 |
| true to $A$ and shared | 23.87 | 27.05 | 28.57 |
| false to $A$ and private | 26.08 | 22.94 | 21.42 |
| false to $A$ and shared | 23.91 | 27.05 | 28.57 |

Table 1: Summary of the constructed datasets (after balancing the training set) and proportion of each class.

(TFxPS), plus three variations with reduced dimensions: Only true/false (TF), only private/shared (PS) and merging true/false on the private cases only (PxTSFS) and iii) control tasks (Hewitt and Liang, 2019) (a) replacing $r$ by a random vector (b) replacing $r$ by a null vector, both only on the training set, to quantify how much information can be extracted from propositions alone during training.

**Evaluation**. Results are evaluated with accuracy on class predictions. To avoid any influence that knowing the position in the dialogue could have (early in the dialogue, propositions have a greater chance of being *private*, and vice versa), we evaluate the results at turn 5 (at which there is a more balanced chance of a fact having been mentioned or not). For the error analysis, we reconstruct complete predicted scoreboards and evaluate incremental aspects: In each column, only one shift from private to shared should occur at the right turn (except for caption propositions, which are always shared) and the true/false status should not change.

**Implementation**. The classifier is implemented with PyTorch (Paszke et al., 2019) and trained with gradient descent using Adam optimizer (Kingma and Ba, 2014) to minimize cross entropy.[6]

## 6 Results

Table 2 presents the accuracy of all models and tasks at turn 5. The performance on the main task is very similar across encoders, with differences lower than 1.5%. $Q$ outperforms $A$ in all models in the main task. While this is expected, since $Q$'s representations must only keep track of the dialogue whereas $A$ must interpret the image, the difference is only marginal.

[6]See Appendix for hyperparameters, model configurations and details on reproducibility. Our code and documentation are available at https://github.com/briemadu/scorekeeping.

| task | | TFxPS | | | TF | | | PS | | | PxTSFS | | |
|---|---|---|---|---|---|---|---|---|---|---|---|---|---|
| | model | (a) | (b) | (c) | (a) | (b) | (c) | (a) | (b) | (c) | (a) | (b) | (c) |
| A | main | 61.80 | 62.37 | 61.31 | 73.05 | 72.50 | 72.41 | 77.29 | 77.31 | 77.13 | 65.57 | 65.49 | 65.83 |
| | random $r$ | 35.25 | 37.52 | 36.60 | 52.25 | 52.01 | 53.17 | 64.59 | 68.52 | 64.07 | 35.46 | 39.22 | 37.48 |
| | null $r$ | 37.43 | 37.19 | 37.42 | 50.65 | 50.65 | 50.67 | 62.79 | 62.85 | 62.66 | 37.36 | 37.51 | 37.35 |
| Q | main | - | - | - | - | - | - | 78.36 | 79.31 | 79.21 | 66.87 | 65.65 | 66.38 |
| | random $r$ | - | - | - | - | - | - | 60.44 | 60.53 | 61.43 | 35.49 | 34.58 | 34.86 |
| | null $r$ | - | - | - | - | - | - | 62.42 | 62.38 | 62.50 | 37.28 | 37.15 | 37.11 |

Table 2: Accuracy on test set at turn 5 (32,360 datapoints) for models (a) RL_DIV, (b) SL, (c) ICCV_RL. TFxPS and TF are not applicable to $Q$ because it has no information to distinguish between what $A$ considers true or false on the private dimension. The hypothesis that results of control tasks do not differ from their corresponding main task is rejected for all cases using paired approximate permutation tests with 1,000 shuffles (p-value$< 0.01$).

For the TF task, the performance on the control tasks is close to random, as expected, but it is higher than random for other tasks. We notice that, while the training dataset is constructed to be balanced in the true/false dimension, information on the private/shared dimension has an inherent bias that is more complex to counterbalance on the training set. Despite the fact that datapoints in the private class do not substantially outnumber the shared class, we observe that each proposition type can have a tendency to occur either early or late in the dialogue (examples in Figure 2), causing them to have an individual skewed distribution towards shared or private at turn 5. This information leak can be used as a shortcut by the classifier.[7] Still, $A$ and $Q$'s representations lead to performances between 8% and 32% higher than the control tasks in all cases.

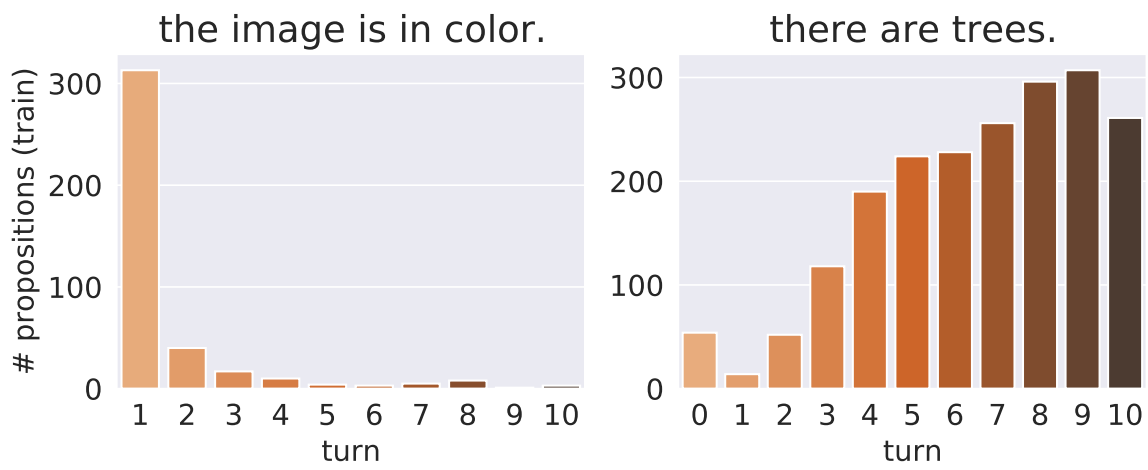

Figure 2: Examples of skewed distributions over dialogue turns which can introduce bias on the private/shared dimension.

**Human Performance**. Table 3 shows the human performance, estimated as the average accuracy of 3 annotators (0.86 Fleiss' $\kappa$ on TFxPS) on a sample of 94 datapoints, each from a different dialogue in the test set (not only at turn 5). We observe that humans agree most of the times on their judgements

| | task | TFxPS | TF | PS | PxTSFS |
|---|---|---|---|---|---|
| | human | 91.84 | 94.32 | 97.51 | 96.09 |
| A | RL_DIV | 52.12 | 65.95 | 74.46 | 65.95 |
| | SL | 50.00 | 72.34 | 73.40 | 68.08 |
| | ICCV_RL | 52.12 | 71.27 | 77.65 | 67.02 |
| Q | RL_DIV | - | - | 75.53 | 62.76 |
| | SL | - | - | 79.78 | 70.21 |
| | ICCV_RL | - | - | 75.53 | 68.08 |

Table 3: Accuracy of human judgement compared to the models on a sample (n=94, not only at turn 5).

and all models perform well below human level.

**Error Analysis**. We conduct an error analysis on $A$, main task, TFxPS. The confusion matrix in Figure 3 shows that it is easier to distinguish between true/false to $A$ in the shared dimension, which can be a sign that dialogue information is more salient in the representations than the image.

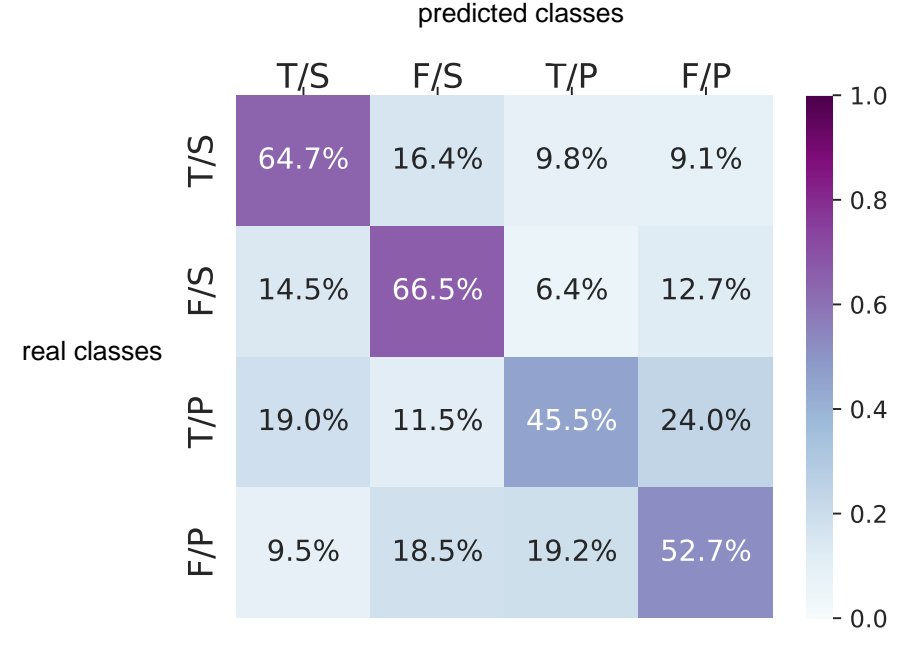

Figure 3: Confusion matrix of predictions at turn 5.

Corrigendum: The axis labels in Figure 3 were swapped but have been corrected in this version.

The accuracy on all datapoints with proposition types that occur on the training set is 67.69, higher than for those that do not, which is 53.11.

When we reconstruct full predicted scoreboards, some qualitative shortcomings become evident. A shift from private to shared is predicted at the correct turn for 60.32% of the propositions but only

[7] As pointed by one of the reviewers, this may not be a shortcoming, since it is how dialogue works and humans are probably also exploiting this.

38.24% shifts *only* at the correct turn. Besides, only 44.50% of the propositions have stable predictions regarding the true/false to $A$ dimension.

Figure 4 shows types of errors in the predictions (the Appendix has more examples). We see the same truth value assigned to opposite propositions, the same proposition classified both as true and false at different turns, as well as an occasional oscillation between private/shared throughout the dialogue. These are indications that, although accuracy per label is generally high, the representations do not seem to always allow incrementally stable and consistent predictions throughout the dialogue.

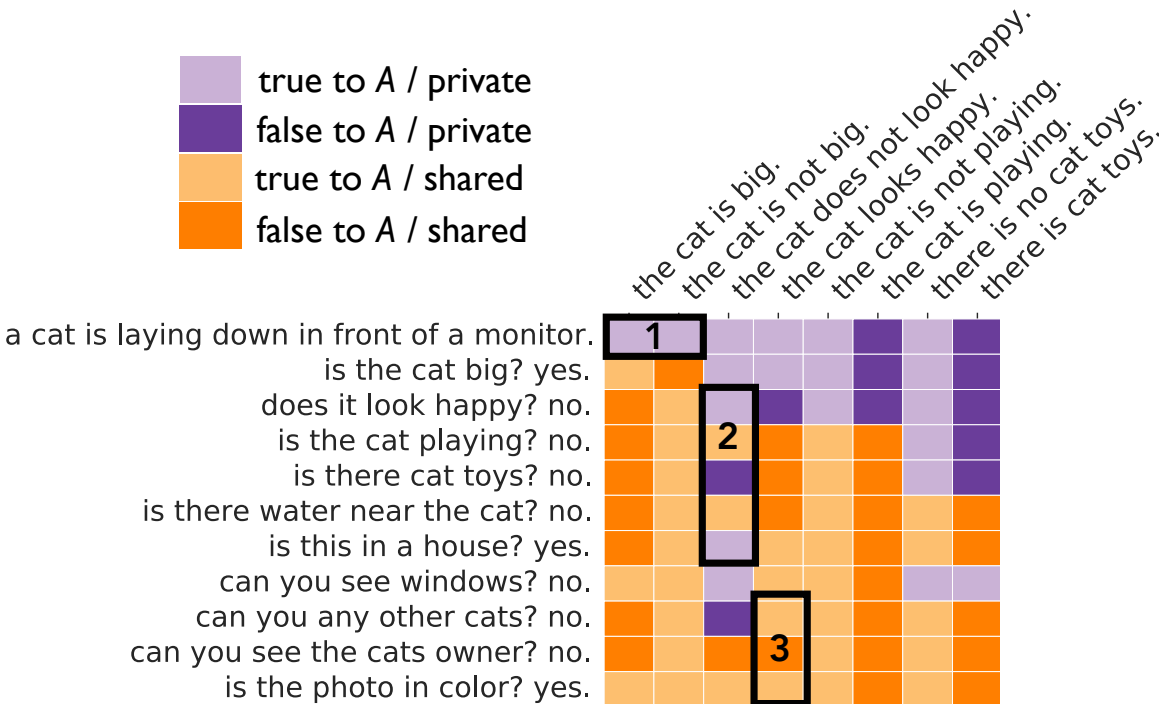

Figure 4: A portion of a predicted scoreboard with some highlighted errors: 1) the same truth value on opposite propositions, 2) oscillation between private and shared, 3) opposite truth values on the same proposition.

## 7 Scope and Limitations

The results on this paper comprise three visual dialogue models trained using a similar setting on the same dataset. The preprocessing steps used by these models replace some tokens by a UNK token and truncate long captions, which prevents some information to become shared as assumed. Further investigation with other models and data is necessary in future research in order to support more general conclusions. The results also rely on the capabilities of the classifier. Although we performed hyperparameter search, the probing classifier does not completely overfit the full training dataset, thus other architectures and hyperparatemeters can be further investigated.

The rule-based generation of propositions has limitations. It cannot generate propositions for all QA pairs and some rules end up not always yielding grammatically valid sentences, for instance because of countable/uncountable nouns, detection of singular/plural forms and mistakes and typos deriving from the dialogues themselves. Besides, spurious patterns deriving from the implemented rules or other confounds and inherent biases (*e.g.* Figure 2) may exist and be predictive of the classes, which could be captured by the probing classifier and influence (likely overestimating) the results. Enforcing a balance on the training set in terms of true/false to $A$ solves one source of bias but causes its distribution to differ from the validation and test set. The test set also has a different distribution because of its varying number of turns.

Finally, while the assumptions proposed in Section 3 are necessary idealizations for using VisDial for this task, they simplify essential aspects of dialogues, *e.g.* the uncertainty about a fact actually being shared, memory limitations and the many kinds of inference that are used in the accommodation of shared knowledge, such as presuppositions, implicatures, entailments and implicit information. Our method cannot capture background knowledge not explicitly stated in dialogue turns.[8]

## 8 Conclusion

We have proposed a novel way to do theory-based evaluation of visual dialogue models. Using diagnostic propositions, we investigated to what degree neural network visual dialogue models incrementally build up representations that are appropriate to do *scorekeeping* of shared commitments throughout a dialogue. The evaluated models trained on VisDial capture part of this process, but not always consistently, possibly because this ability is not an elementary component of the training regime. The relatively impoverished nature of the original task in terms of *coordination phenomena* can also limit the capability of models to build good dialogue representations (Schlangen, 2019). Future work should extend the evaluation to other models and reflect on how better and ecologically valid diagnostic datasets for visual dialogues can be constructed.

## 9 Ethical Considerations

Propositions are direct manipulations of QA pairs and thus reflect the subjective judgments of VisDial crowdworkers. Therefore, they are not *per se* necessarily *true* or *false* with respect to the image, but with respect to $A$'s interpretation expressed as answers. Inappropriate content on images, captions and dialogues can be replicated by the rule-based proposition generation. To try to remedy this, we filtered out dialogues containing words that could be used for sensitive content. Despite our efforts, we cannot guarantee that we could remove everything, given the size of the dataset and the inherent bias of how humans interpret images. As a result, the only purpose of the propositions is performing the evaluation as proposed here.

[8]We thank the reviewers for pointing out some of the limitations discussed in this section.

## Acknowledgements

We are thankful to the anonymous reviewers for their feedback and suggestions, to Wencke Liermann for implementing the interface for the human evaluation and to the student assistants of the Computational Linguistics Lab who contributed on the experiment.

## References

Shubham Agarwal, Trung Bui, Joon-Young Lee, Ioannis Konstas, and Verena Rieser. 2020. History for visual dialog: Do we really need it? In *Proceedings of the 58th Annual Meeting of the Association for Computational Linguistics*, pages 8182–8197, Online. Association for Computational Linguistics.

Yonatan Belinkov and James Glass. 2019. Analysis methods in neural language processing: A survey. *Transactions of the Association for Computational Linguistics*, 7:49–72.

Luciana Benotti and Patrick Blackburn. 2021. Grounding as a collaborative process. In *Proceedings of the 16th Conference of the European Chapter of the Association for Computational Linguistics: Main Volume*, pages 515–531, Online. Association for Computational Linguistics.

Yonatan Bitton, Gabriel Stanovsky, Roy Schwartz, and Michael Elhadad. 2021. Automatic generation of contrast sets from scene graphs: Probing the compositional consistency of GQA. In *Proceedings of the 2021 Conference of the North American Chapter of the Association for Computational Linguistics: Human Language Technologies*, pages 94–105, Online. Association for Computational Linguistics.

Herbert H Clark and Susan E Brennan. 1991. Grounding in communication. In *Perspectives on socially shared cognition.*, pages 127–149. American Psychological Association.

Abhishek Das, Satwik Kottur, Khushi Gupta, Avi Singh, Deshraj Yadav, José MF Moura, Devi Parikh, and Dhruv Batra. 2017a. Visual dialog. In *Proceedings of the IEEE Conference on Computer Vision and Pattern Recognition*, pages 326–335.

Abhishek Das, Satwik Kottur, José MF Moura, Stefan Lee, and Dhruv Batra. 2017b. Learning cooperative visual dialog agents with deep reinforcement learning. In *Proceedings of the IEEE international conference on computer vision*, pages 2951–2960.

Dorottya Demszky, Kelvin Guu, and Percy Liang. 2018. Transforming question answering datasets into natural language inference datasets. *arXiv preprint arXiv:1809.02922*.

Nouha Dziri, Ehsan Kamalloo, Kory Mathewson, and Osmar Zaiane. 2019. Evaluating coherence in dialogue systems using entailment. In *Proceedings of the 2019 Conference of the North American Chapter of the Association for Computational Linguistics: Human Language Technologies, Volume 1 (Long and Short Papers)*, pages 3806–3812, Minneapolis, Minnesota. Association for Computational Linguistics.

Fabian Galetzka, Jewgeni Rose, David Schlangen, and Jens Lehmann. 2021. Space efficient context encoding for non-task-oriented dialogue generation with graph attention transformer. In *Proceedings of the 59th Annual Meeting of the Association for Computational Linguistics and the 11th International Joint Conference on Natural Language Processing (Volume 1: Long Papers)*, pages 7028–7041, Online. Association for Computational Linguistics.

Jonathan Ginzburg. 2012. *The Interactive Stance.* Chapter 4: Basic Interaction in Dialogue. Oxford University Press.

Claudio Greco, Alberto Testoni, and Raffaella Bernardi. 2020a. Grounding dialogue history: Strengths and weaknesses of pre-trained transformers. In *International Conference of the Italian Association for Artificial Intelligence*, pages 263–279. Springer.

Claudio Greco, Alberto Testoni, and Raffaella Bernardi. 2020b. Which turn do neural models exploit the most to solve GuessWhat? Diving into the dialogue history encoding in transformers and lstms. In *Proceedings of the 4th Workshop on Natural Language for Artificial Intelligence (NL4AI 2020) co-located with the 19th International Conference of the Italian Association for Artificial Intelligence (AI*IA 2020), Anywhere, November 25th-27th, 2020*, volume 2735 of *CEUR Workshop Proceedings*, pages 29–43. CEUR-WS.org.

John Hewitt and Percy Liang. 2019. Designing and interpreting probes with control tasks. In *Proceedings of the 2019 Conference on Empirical Methods in Natural Language Processing and the 9th International Joint Conference on Natural Language Processing (EMNLP-IJCNLP)*, pages 2733–2743, Hong Kong, China. Association for Computational Linguistics.

Dieuwke Hupkes, Sanne Bouwmeester, and Raquel Fernández. 2018a. Analysing the potential of seq-to-seq models for incremental interpretation in task-oriented dialogue. In *Proceedings of the 2018 EMNLP Workshop BlackboxNLP: Analyzing and Interpreting Neural Networks for NLP*, pages 165–174, Brussels, Belgium. Association for Computational Linguistics.

Dieuwke Hupkes, Sara Veldhoen, and Willem Zuidema. 2018b. Visualisation and diagnostic classifiers' reveal how recurrent and recursive neural networks process hierarchical structure. *Journal of Artificial Intelligence Research*, 61:907–926.

Justin Johnson, Bharath Hariharan, Laurens van der Maaten, Li Fei-Fei, C. Lawrence Zitnick, and Ross Girshick. 2017. Clevr: A diagnostic dataset for compositional language and elementary visual reasoning. In *Proceedings of the IEEE Conference on Computer Vision and Pattern Recognition (CVPR)*.

Diederik P Kingma and Jimmy Ba. 2014. Adam: A method for stochastic optimization. *arXiv preprint arXiv:1412.6980*.

Kenton Lee, Luheng He, and Luke Zettlemoyer. 2018. Higher-order coreference resolution with coarse-to-fine inference. In *Proceedings of the 2018 Conference of the North American Chapter of the Association for Computational Linguistics: Human Language Technologies, Volume 2 (Short Papers)*, pages 687–692, New Orleans, Louisiana. Association for Computational Linguistics.

David Lewis. 1979. Scorekeeping in a language game. In *Semantics from different points of view*, pages 172–187. Springer.

Tsung-Yi Lin, Michael Maire, Serge Belongie, James Hays, Pietro Perona, Deva Ramanan, Piotr Dollár, and C Lawrence Zitnick. 2014. Microsoft coco: Common objects in context. In *European conference on computer vision*, pages 740–755. Springer.

Sharid Loáiciga, Simon Dobnik, and David Schlangen. 2021. Reference and coreference in situated dialogue. In *Proceedings of the Second Workshop on Advances in Language and Vision Research*, pages 39–44, Online. Association for Computational Linguistics.

Vishvak Murahari, Prithvijit Chattopadhyay, Dhruv Batra, Devi Parikh, and Abhishek Das. 2019. Improving generative visual dialog by answering diverse questions. In *Proceedings of the 2019 Conference on Empirical Methods in Natural Language Processing and the 9th International Joint Conference on Natural Language Processing (EMNLP-IJCNLP)*, pages 1449–1454.

Prasanna Parthasarathi, Joelle Pineau, and Sarath Chandar. 2020. How to evaluate your dialogue system: Probe tasks as an alternative for token-level evaluation metrics. *arXiv preprint arXiv:2008.10427*.

Adam Paszke, Sam Gross, Francisco Massa, Adam Lerer, James Bradbury, Gregory Chanan, Trevor Killeen, Zeming Lin, Natalia Gimelshein, Luca Antiga, Alban Desmaison, Andreas Kopf, Edward Yang, Zachary DeVito, Martin Raison, Alykhan Tejani, Sasank Chilamkurthy, Benoit Steiner, Lu Fang, Junjie Bai, and Soumith Chintala. 2019. Pytorch: An imperative style, high-performance deep learning library. In H. Wallach, H. Larochelle, A. Beygelzimer, F. d'Alché-Buc, E. Fox, and R. Garnett, editors, *Advances in Neural Information Processing Systems 32*, pages 8024–8035. Curran Associates, Inc.

Nils Reimers and Iryna Gurevych. 2019. Sentence-BERT: Sentence embeddings using Siamese BERT-networks. In *Proceedings of the 2019 Conference on Empirical Methods in Natural Language Processing and the 9th International Joint Conference on Natural Language Processing (EMNLP-IJCNLP)*, pages 3982–3992, Hong Kong, China. Association for Computational Linguistics.

Marco Tulio Ribeiro, Carlos Guestrin, and Sameer Singh. 2019. Are red roses red? evaluating consistency of question-answering models. In *Proceedings of the 57th Annual Meeting of the Association for Computational Linguistics*, pages 6174–6184, Florence, Italy. Association for Computational Linguistics.

Marco Tulio Ribeiro, Sameer Singh, and Carlos Guestrin. 2018. Semantically equivalent adversarial rules for debugging NLP models. In *Proceedings of the 56th Annual Meeting of the Association for Computational Linguistics (Volume 1: Long Papers)*, pages 856–865, Melbourne, Australia. Association for Computational Linguistics.

Abdelrhman Saleh, Tovly Deutsch, Stephen Casper, Yonatan Belinkov, and Stuart Shieber. 2020. Probing neural dialog models for conversational understanding. In *Proceedings of the 2nd Workshop on Natural Language Processing for Conversational AI*, pages 132–143, Online. Association for Computational Linguistics.

Chinnadhurai Sankar, Sandeep Subramanian, Chris Pal, Sarath Chandar, and Yoshua Bengio. 2019. Do neural dialog systems use the conversation history effectively? an empirical study. In *Proceedings of the 57th Annual Meeting of the Association for Computational Linguistics*, pages 32–37, Florence, Italy. Association for Computational Linguistics.

David Schlangen. 2019. Grounded agreement games: Emphasizing conversational grounding in visual dialogue settings. *arXiv cs.CL preprint arXiv:1908.11279*.

Iulian Serban, Alessandro Sordoni, Yoshua Bengio, Aaron Courville, and Joelle Pineau. 2016. Building end-to-end dialogue systems using generative hierarchical neural network models. In *Proceedings of the AAAI Conference on Artificial Intelligence*, volume 30.

Ravi Shekhar, Sandro Pezzelle, Yauhen Klimovich, Aurélie Herbelot, Moin Nabi, Enver Sangineto, and Raffaella Bernardi. 2017. FOIL it! find one mismatch between image and language caption. In *Proceedings of the 55th Annual Meeting of the Association for Computational Linguistics (Volume 1: Long Papers)*, pages 255–265, Vancouver, Canada. Association for Computational Linguistics.

Ravi Shekhar, Ece Takmaz, Raquel Fernández, and Raffaella Bernardi. 2019a. Evaluating the representational hub of language and vision models. In *Proceedings of the 13th International Conference on Computational Semantics - Long Papers*, pages 211–222, Gothenburg, Sweden. Association for Computational Linguistics.

Ravi Shekhar, Aashish Venkatesh, Tim Baumgärtner, Elia Bruni, Barbara Plank, Raffaella Bernardi, and Raquel Fernández. 2019b. Beyond task success: A closer look at jointly learning to see, ask, and GuessWhat. In *Proceedings of the 2019 Conference of the North American Chapter of the Association for Computational Linguistics: Human Language Technologies, Volume 1 (Long and Short Papers)*, pages 2578–2587, Minneapolis, Minnesota. Association for Computational Linguistics.

Alessandro Sordoni, Michel Galley, Michael Auli, Chris Brockett, Yangfeng Ji, Margaret Mitchell, Jian-Yun Nie, Jianfeng Gao, and Bill Dolan. 2015. A neural network approach to context-sensitive generation of conversational responses. In *Proceedings of the 2015 Conference of the North American Chapter of the Association for Computational Linguistics: Human Language Technologies*, pages 196–205, Denver, Colorado. Association for Computational Linguistics.

Oriol Vinyals and Quoc Le. 2015. A neural conversational model. In *ICML Deep Learning Workshop*.

Sean Welleck, Jason Weston, Arthur Szlam, and Kyunghyun Cho. 2019. Dialogue natural language inference. In *Proceedings of the 57th Annual Meeting of the Association for Computational Linguistics*, pages 3731–3741, Florence, Italy. Association for Computational Linguistics.

Chien-Sheng Wu and Caiming Xiong. 2020. Probing task-oriented dialogue representation from language models. In *Proceedings of the 2020 Conference on Empirical Methods in Natural Language Processing (EMNLP)*, pages 5036–5051, Online. Association for Computational Linguistics.

Chen Zhang and Joyce Chai. 2009. What do we know about conversation participants: Experiments on conversation entailment. In *Proceedings of the SIGDIAL 2009 Conference*, pages 206–215, London, UK. Association for Computational Linguistics.

Chen Zhang and Joyce Chai. 2010. Towards conversation entailment: An empirical investigation. In *Proceedings of the 2010 Conference on Empirical Methods in Natural Language Processing*, pages 756–766, Cambridge, MA. Association for Computational Linguistics.

# Appendix

## A Generating Propositions and Constructing the Datasets

This section presents details about the procedure to turn QA pairs from the VisDial dataset[9] into propositions.

**Solving Pronouns**. Coreference resolution is specially challenging on visual dialogues, as discussed in Loáiciga et al. (2021). Despite the limitations, we used the model proposed in Lee et al. (2018) to replace pronouns (those that were detected and solved) by their corresponding entity as follows:

1. Merged caption and QA pairs into a single string.
2. Passed string to coreference resolution model to get coreference clusters.[10]
3. Assumed that the first element in the cluster was the entity (its first mention).
4. For each dialogue, checked which questions and answers contained pronouns of interest (he, she, it, they, his, her, its, their, him, them, hers, theirs, this, that, these, those) and replaced them with their corresponding cluster entity, if detected. Assumed the pronoun her was always possessive.
5. If the entity comprised more than N=5 tokens, we did not replace it (because entities spanning over many tokens are very likely to be long portions of the caption that result in wrong propositions).
6. With postprocessing steps, put string back into VisDial format.

On average, 2.24 pronouns were replaced per dialogue on the training set, 2.43 on the validation set and 1.15 on the test set.

**Generating Propositions**. Automatic generation of diagnostic datasets or adversarial examples via programmatic manipulation rules or templates is a usual step in probing studies, *e.g.* Johnson et al. (2017), Shekhar et al. (2017), Ribeiro et al. (2018) and Bitton et al. (2021). The main steps to turn QA pairs into propositions were to some extent based on Ribeiro et al. (2019) and Demszky et al. (2018). We analysed common patterns of questions and answers on VisDial and implemented 34 rules that create entailments and contradictions. Some rules are lexical (*e.g.* questions starting with '*what color is*' and whose answer has a color name) and others depend on POS tag patterns extracted using SpaCy v.3.0.5.[11] Most rules work for polar questions, some work for other types of questions. We noticed that some images and dialogues on VisDial contain inappropriate content. To avoid replicating this on the propositions, we filtered out dialogues that contain words that may be sensitive (see code documentation for details). Propositions were then generated as follows:

1. Parsed the caption to extract nouns and adjectives and generated caption propositions.
2. For each turn, checked whether it matched a manipulation rule.
3. Every rule, when they were applied, generated a direct entailment and a direct contradiction (negation of the entailment).
4. Propositions that contained pronouns (for cases in which coreference resolution did not work), except for *it*, or that were too long (more then 15 tokens) were excluded.

[9] Available at https://visualdialog.org/

[10] Implementation by AllenNLP, version 2.1.0, at https://demo.allennlp.org/coreference-resolution with their pretrained model coref-spanbert-large-2021.03.10.

The code documentation has a more detailed description of the rules. The next sections present details of the resulting proposition sets. Note that the number of dialogues in each set is smaller than in the VisDial original splits, because some were filtered out and others had no propositions.

Propositions have four attributes: i) kind of manipulation rule; ii) dialogue and turn from which it derives; iii) a true/false status with respect to what $A$ thinks about the image; iv) the polarity (positive/negative) of the answer, if applicable.

**Downsampling and de-biasing**. We noticed that the proportion of caption propositions was much larger than propositions deriving from other turns, which would cause a considerable imbalance towards facts that are always shared in the scoreboard. Therefore, we sampled 15% of the caption pairs (entailment and contradiction) on all datasets to make the distribution over manipulated turns be closer to uniform.

Furthermore, in preliminary experiments we observed that propositions could give away information on the true/false to $A$ status. For instance, '*there is a zebra.*' can appear very often as an entailment (on the many photos showing zebras) but rarely as a contradiction (dialogues where $Q$ spontaneously asks '*is there a zebra?*' and the answer is '*no*'). Besides, on rules that manipulate questions that are not polar (*what color is the dog? black.*), negation is always a contradiction. So the classifier could make predictions based on the lexical form alone. To counter this bias, we constructed a balanced training dataset by sampling from the original set while making sure that, for each $p$ that $A$ established to be true with respect to an image/dialogue, we also included an equal $p$ paired with an image/dialogue in which it is established to be false. While this procedure reduced the size of the training set, we ensured that predictions on the true/false dimension would need to use the dialogue representations. We also limited the number of $p$ of the same kind to 2,000 (1,000 as entailment, 1,000 as contradiction), to avoid having very common propositions like '*the photo is in color*' or '*it is sunny*' occurring too often.

**Datasets used in the experiments.** The following paragraphs discuss the final datasets used in the experiments (*i.e.* after downsampling captions and balancing the training set). The frequency over which turn was manipulated is shown in Figure 5. Although there is an imbalance towards later turns on the training set, the proportion of private/shared classes at turn 5 is relatively balanced (around 44.5/55.5), partially due to the fact that, at the last turn, no proposition is assigned a private class. Figure 6 shows the frequency of the number of turns that have been turned into propositions in a dialogue. Table 4 show the proportion of each type of proposition on the datasets. The training set has less propositions that do not derive from polar questions due to the balancing.

The propositions, paired to dialogue representations on each dialogue turn, with the class assigned to each tuple can be seen as a layer of annotation which is not predicted but constructed.

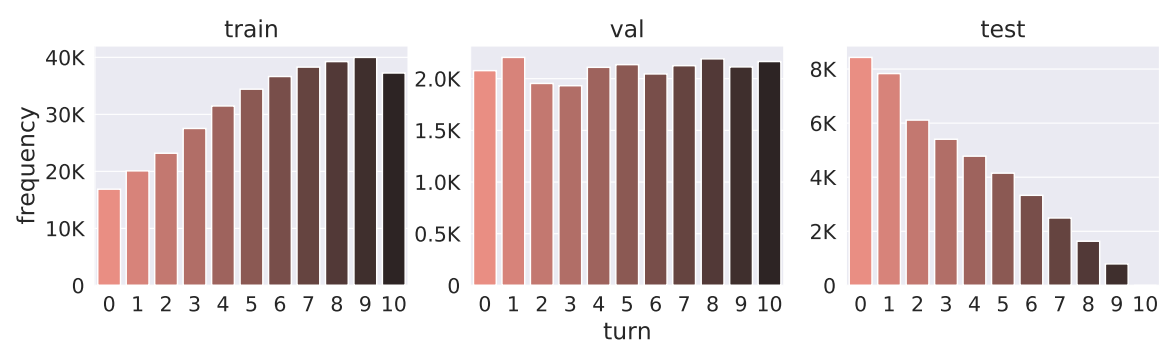

Figure 5: Distribution over manipulated turns. The test set has a different distribution because it has incomplete dialogues with varying length.

37.20% of the validation proposition types and 31.58% of the test proposition types appear among the training propositions. 82.68% of the validation propositions and 79.63% of the test propositions occur in only one dialogue. On average, a proposition appears in 12.77 dialogues in the training set, 1.91 dialogues in the validation set and 2.34 dialogues in the test set. 72.73% of the word types in the validation set and 63.00% of the word types in the test set occur in the training set.

[11] https://spacy.io/

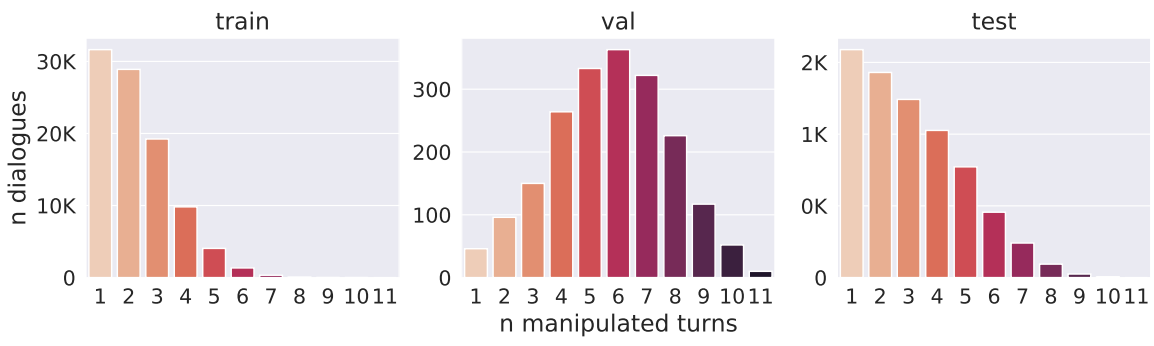

Figure 6: Number of manipulated turns per dialogue.

| | train | valid | test |
|---|---|---|---|
| true to $A$ | 50.00 | 50.00 | 50.00 |
| false to $A$ | 50.00 | 50.00 | 50.00 |
| polar $q$, positive $a$ | 43.17 | 32.73 | 31.35 |
| polar $q$, negative $a$ | 49.97 | 39.09 | 31.44 |
| other $q$ | 6.84 | 28.16 | 37.19 |

Table 4: Proportion (%) of each type of proposition.

**Examples**. Figure 10 shows dialogues from the training set and the propositions generated for each turn, after downsampling the caption propositions (but before balancing). Propositions can inherit grammatical or spelling problems from the dialogues themselves. Figure 1 in the main section contains all propositions, before downsampling.

**Collecting dialogue representations**. To collect the dialogue state representations, we adapted the original *train.py* and *evaluate.py* scripts.[12] To get the representation at turn 10 for $A$, we needed to feed a dummy next question made of the start and the end symbols with a question mark token in between.

**Human Judgement**. We randomly sampled 100 dialogues and one proposition on each of them.[13] Then we sampled a random turn up to which the corresponding dialogue would be shown. The annotators were non-native English speakers who worked as student assistants at the Computational Linguistics Lab of the University of Potsdam. The task was explained to the annotators verbally and then again in written form at the beginning of the annotation. All participants saw the same datapoints at a different random order, presented in a setting as shown in Figure 7, and had to select one of the four alternatives (which correspond to the main task TFxPS).

[12] https://github.com/vmurahari3/visdial-diversity

[13] 6 datapoints were later excluded due to a technical mismatch after refactoring.

| | train | valid | test |
|---|---|---|---|
| manipulation rule types | 34 | 34 | 34 |
| avr. manipulated turns per dialogue | 2.28 | 5.72 | 3.13 |
| min. propositions per dialogue | 1 | 2 | 2 |
| max. propositions per dialogue | 16 | 26 | 22 |

Table 5: Details of the proposition sets (after downsampling and balancing).

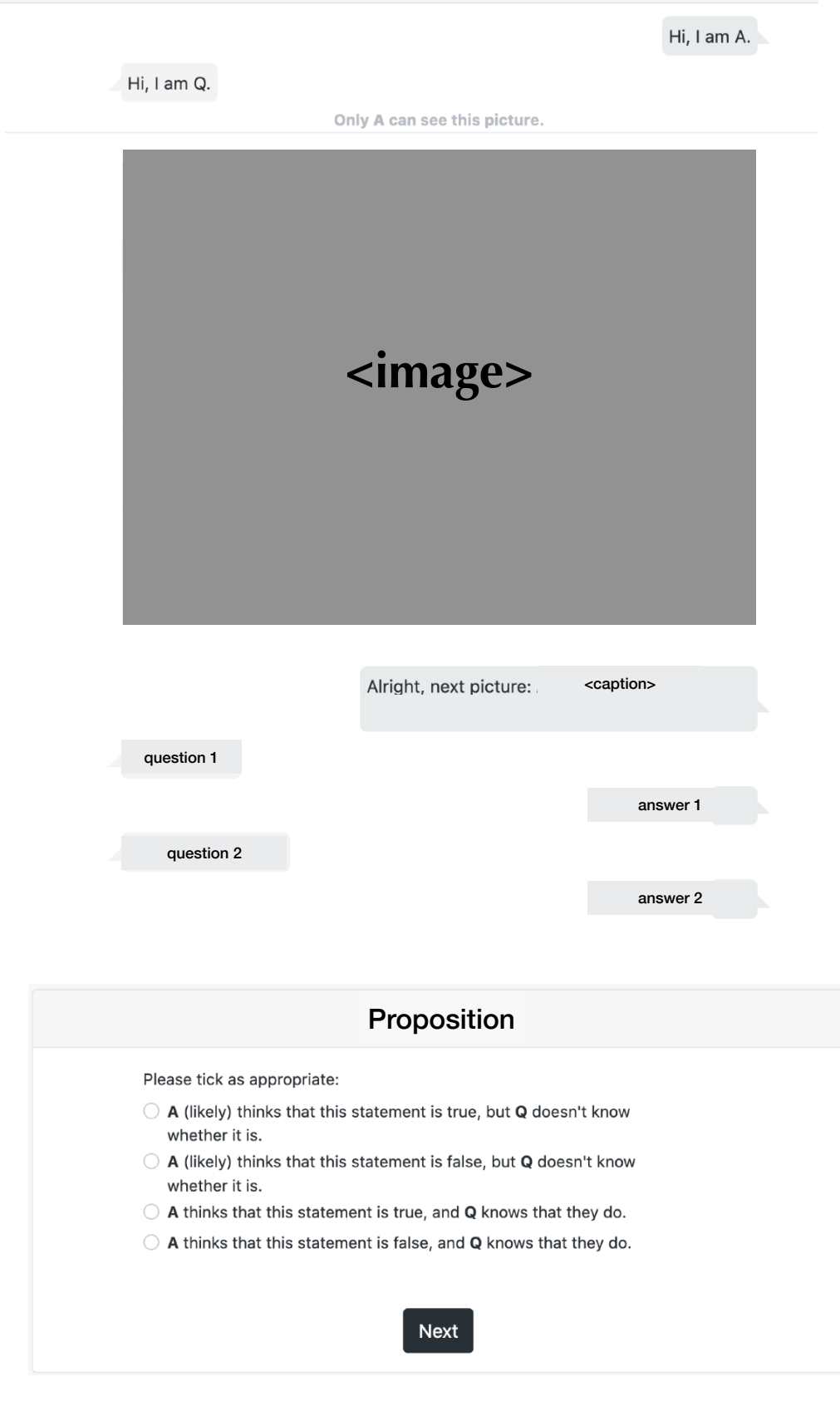

Figure 7: How the task was presented for the annotators.

## B Reproducibility

In this section, we present further details of the implementation and additional results to support reproducibility. More information can also be found in the code documentation.

**Hyperparameters**. We used comet.ml's[14] implementation of the Bayes algorithm for hyperpa-rameter search on $A$, main task, TFxPS, RL_DIV, aiming at maximizing accuracy on the validation set, as well as some manual selections. The (non-exhaustive) search space is shown in Table 6. The optimal configuration was then used in all experiments, with a maximum of 30 epochs and no early-stopping. A preliminary test with an even larger hidden dimension showed a very minor improvement. For each experiment, we used the configuration that led to the best performance on the validation set to get results on the test set. Each experiment took between 50 and 60 minutes.

[14] www.comet.ml

The sentence encoder models listed on Table 6 are available at HuggingFace's Model Hub.[15]

**Classifier architecture**. The neural network was implemented using Pytorch 1.7.1. The proposition embeddings have 768 dimensions and the dialogue context embeddings have 512 dimensions. We used a sequential model from PyTorch with the following layers and dimensions:[16]

1. linear layer (in features=768+512, out features=1024, bias=True)
2. sigmoid function
3. dropout layer (p=0.1)
4. linear layer (in features=1024, out features=n labels in {2,3,4}, bias=True)
5. softmax function + cross entropy loss

The models have 1,315,844, 1,314,819 and 1,313,794 trainable parameters for the classification tasks with 4, 3 and 2 labels, respectively.

**Infrastructure**. The operating system used to run experiments was Linux, release 5.4.0-99-generic, processor x86_64. We had two GPUs available (NVIDIA GeForce GTX 1080 Ti), but each individual experiment used only one of them.

## C Detailed Results

Table 7 shows the overall accuracy on all datapoints (comprising all turns in the test set). Table 8 and Table 9 show all results on the validation set.

On Figure 8 we split the accuracy per type of proposition. Propositions that derive from negative facts about the image (*'is there a dog? no.'*) seem to be harder than positive ones when they derive from earlier turns, but they are easier to correctly classify when they derive from later turns. Propositions deriving from questions that are not polar are harder (which may be a consequence of the balanced dataset selection that results in few propositions of this type for training). We also see that propositions derived from manipulating later turns are, in general, harder to classify.

When we consider each row of the scoreboard (representing the scoreboard at a given turn), we can inspect how accuracy evolves over turns, illustrated in Figure 9.

For the error analysis on captions, a right shift from private to shared means that the class at turn 0 is shared. Shifting only at the right turn means that it starts as shared and does not shift at any turn.

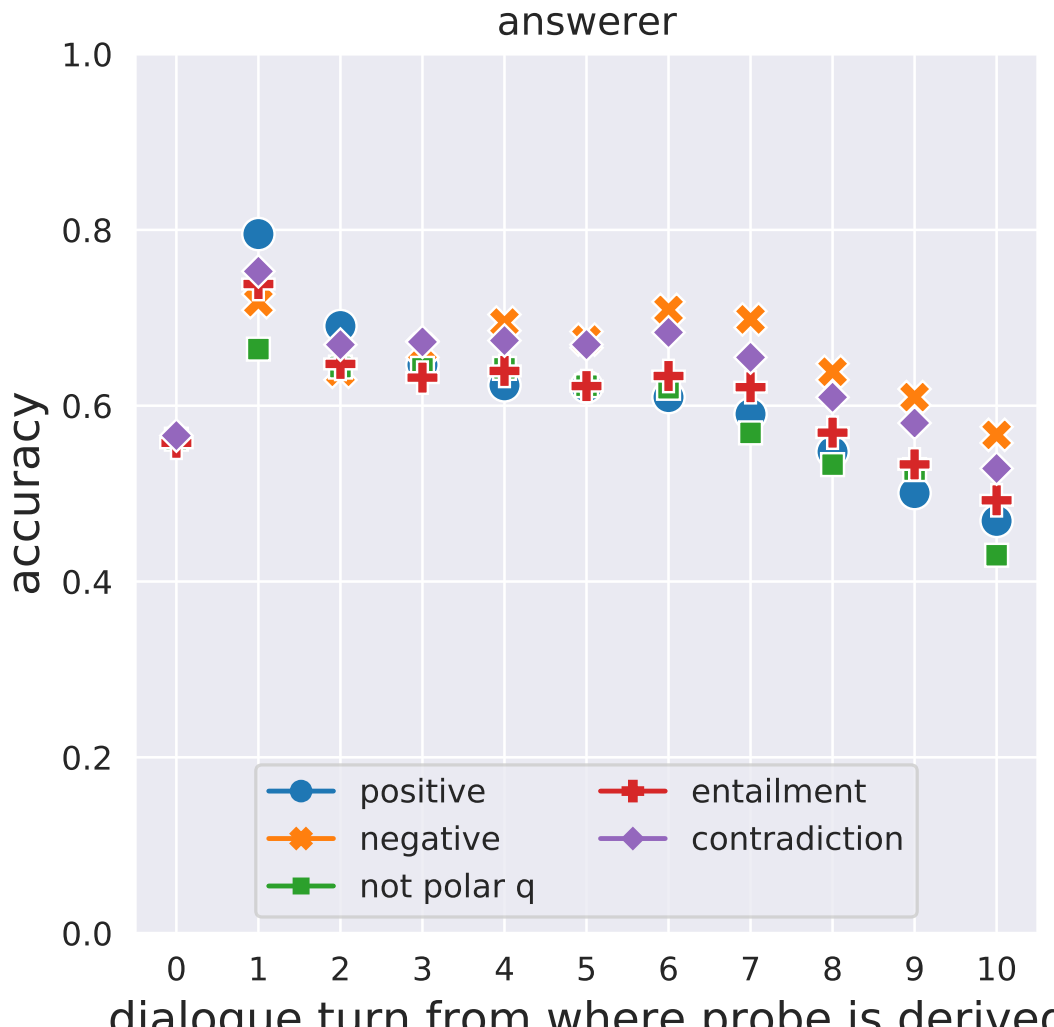

Figure 8: Accuracy per type of proposition ($A$, main, TFxPS, RL_DIV).

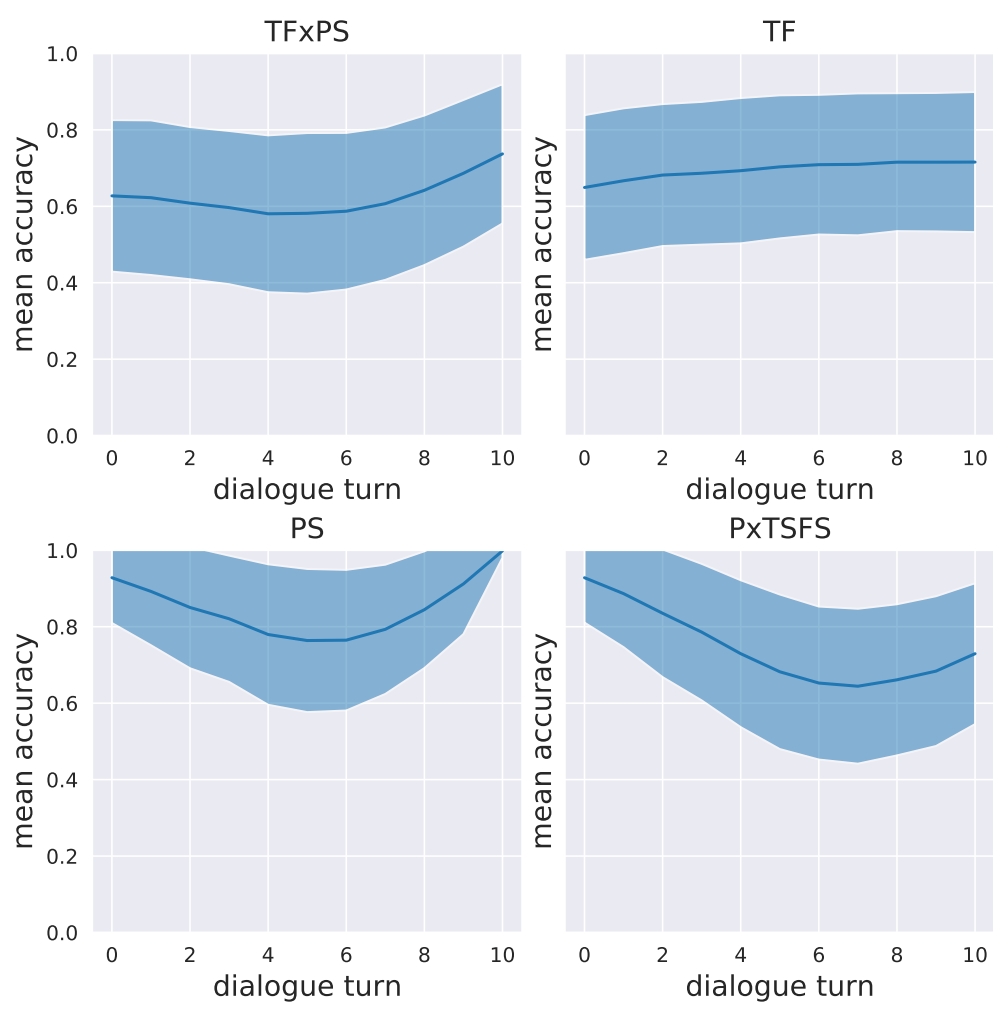

Figure 9: Mean accuracy on dialogue level over turns ($A$, main, TFxPS, RL_DIV).

[15] https://huggingface.co/sentence-transformers

[16] During development, we also experimented with a shallow version, which did not perform very well, and a version with more layers, whose performance gain was not substantial.

| **Hyperparameter** | **Values** | **Selected** |
|---:|---|---|
| batch size | 64, 128, 256, 512 | 512 |
| clipping | 0.0, 0.25, 0.5, 1, 5 | 1 |
| dropout | 0.0, 0.1, 0.3, 0.5 | 0.1 |
| hidden dimension | 64, 128, 256, 512, 1024 | 1024 |
| learning rate | 1e-5, 1e-3, 3e-5, 3e-3, 1e-2 | 0.001 |
| random seed | 2204, 10, 142, 54321 | 54321 |
| sentence encoder | stsb-bert-base, paraphrase-mpnet-base-v2, nli-roberta-base-v2, stsb-roberta-base-v2 | paraphrase-mpnet-base-v2 |

Table 6: Hyperparameters tried in the (non-exhaustive) search and selected hyperparameters used in all final experiments.

| task | | TFxPS | | | TF | | | PS | | | PxTSFS | | |
|---|---|---|---|---|---|---|---|---|---|---|---|---|---|
| model | | (a) | (b) | (c) | (a) | (b) | (c) | (a) | (b) | (c) | (a) | (b) | (c) |
| A | main | 62.04 | 62.33 | 61.78 | 71.02 | 70.92 | 70.79 | 80.94 | 81.24 | 80.79 | 73.06 | 73.36 | 73.47 |
| | random $r$ | 35.10 | 35.56 | 35.12 | 52.48 | 51.82 | 53.17 | 60.35 | 60.65 | 60.46 | 47.95 | 48.65 | 48.62 |
| | null $r$ | 37.66 | 37.52 | 37.71 | 50.61 | 50.60 | 50.61 | 60.25 | 60.24 | 60.21 | 50.64 | 50.86 | 50.62 |
| Q | main | - | - | - | - | - | - | 82.02 | 83.15 | 83.06 | 74.35 | 73.90 | 74.42 |
| | random $r$ | - | - | - | - | - | - | 59.00 | 59.75 | 60.06 | 48.80 | 48.32 | 48.49 |
| | null $r$ | - | - | - | - | - | - | 60.18 | 60.13 | 60.15 | 50.64 | 50.56 | 50.53 |

Table 7: Accuracy on the test set (all turns) for models (a) RL_DIV, (b) SL, (c) ICCV_RL.

| task | | TFxPS | | | TF | | | PS | | | PxTSFS | | |
|---|---|---|---|---|---|---|---|---|---|---|---|---|---|
| model | | (a) | (b) | (c) | (a) | (b) | (c) | (a) | (b) | (c) | (a) | (b) | (c) |
| A | main | 57.97 | 58.03 | 57.31 | 70.32 | 70.45 | 70.35 | 76.31 | 77.17 | 75.97 | 68.13 | 69.15 | 68.41 |
| | random $r$ | 33.48 | 35.76 | 35.53 | 52.45 | 52.93 | 53.69 | 62.09 | 61.85 | 58.51 | 51.14 | 50.72 | 49.90 |
| | null $r$ | 37.44 | 37.39 | 37.55 | 50.75 | 50.75 | 50.75 | 63.94 | 63.91 | 63.92 | 53.05 | 52.95 | 53.10 |
| Q | main | - | - | - | - | - | - | 78.49 | 79.74 | 79.22 | 71.62 | 71.37 | 71.28 |
| | random $r$ | - | - | - | - | - | - | 62.30 | 60.80 | 61.16 | 52.12 | 51.58 | 51.69 |
| | null $r$ | - | - | - | - | - | - | 63.89 | 63.82 | 63.86 | 53.17 | 52.98 | 52.99 |

Table 8: Accuracy on the validation set (turn 5) for models (a) RL_DIV, (b) SL, (c) ICCV_RL.

| task | | TFxPS | | | TF | | | PS | | | PxTSFS | | |
|---|---|---|---|---|---|---|---|---|---|---|---|---|---|
| model | | (a) | (b) | (c) | (a) | (b) | (c) | (a) | (b) | (c) | (a) | (b) | (c) |
| A | main | 62.46 | 62.55 | 62.30 | 69.59 | 69.83 | 69.52 | 85.00 | 85.34 | 84.82 | 74.74 | 75.13 | 74.97 |
| | random $r$ | 33.52 | 33.86 | 33.54 | 52.42 | 52.85 | 53.55 | 59.54 | 59.64 | 59.88 | 49.53 | 49.55 | 50.02 |
| | null $r$ | 34.84 | 34.75 | 34.88 | 50.74 | 50.74 | 50.74 | 59.75 | 59.73 | 59.71 | 51.14 | 51.01 | 51.13 |
| Q | main | - | - | - | - | - | - | 85.33 | 86.23 | 86.37 | 76.23 | 75.79 | 76.15 |
| | random $r$ | - | - | - | - | - | - | 58.70 | 60.43 | 60.45 | 50.01 | 49.88 | 50.02 |
| | null $r$ | - | - | - | - | - | - | 59.68 | 59.63 | 59.63 | 51.25 | 51.16 | 51.16 |

Table 9: Accuracy on the validation set (all turns) for models (a) RL_DIV, (b) SL, (c) ICCV_RL.

**a dog that is looking at a herd of sheep.**
none
**are there any people? no.**
there are no people.
there are people.
**what color is the dog? whitish tan.**
the dog is tan.
the dog is not tan.
**is this in color? yes.**
the image is in color.
the image is not in color.
**is this a large field? very large.**
none
**is there tall grass? no.**
there is no tall grass.
there is tall grass.
**is it sunny? a little.**
none
**can you see a fence? no fences.**
one cannot see any fence.
one can see a fence.
**are there trees? 0.**
there are no trees.
there are trees.
**can you see mountains? i see a hillside.**
none
**any buildings? no buildings at all.**
there are no buildings.
there are buildings.

**this is a white kitchen with a window.**
none
**do you see a stove? yes.**
one can see a stove.
one cannot see any stove.
**what color is the stove? white and black.**
the stove is white and black.
the stove is not white and black.
**do you see a sink? yes.**
one can see a sink.
one cannot see any sink.
**can you see the fridge? no.**
one cannot see any fridge.
one can see a fridge.
**do the window have any curtains? no curtains.**
the window do not have any curtains.
the window have any curtains.
**do you see a dishwasher? no.**
one cannot see any dishwasher.
one can see a dishwasher.
**do you see any blinds? no blinds.**
one cannot see any blinds.
one can see blinds.
**any pictures on the wall? 0.**
there are no pictures on the wall.
there are pictures on the wall.
**do you see any people? no people are in the room.**
one cannot see any people.
one can see people.
**what color is the floors? grey.**
the floors is grey.
the floors is not grey.

**a serving of dessert that includes various berries.**
none
**does this food look appetizing? no.**
none
**is veggies on dish? nope just fruit.**
none
**do you see apples? no apples.**
one cannot see any apples.
one can see apples.
**do you see grapes? no gapes at all.**
one cannot see any grapes.
one can see grapes.
**what is main fruit on dish? strawberries and blueberries.**
none
**do strawberries still have green on them? yes it does.**
none
**are blueberries large? no small and smashed.**
the blueberries are not large.
the blueberries are large.
**can you tell what color plate is? it is white bowl.**
none
**can you tell color of table? no,.**
none
**do you see people? no.**
one cannot see any people.
one can see people.

**a black cat laying in the sun on a green bench.**
one can see a black cat.
one cannot see a black cat.
**is the bench chipped? no it's not.**
the bench is not chipped.
the bench is chipped.
**is it wood or metal? it looks metal to me.**
none
**is the cat sleep? no i see the eye to be open.**
the cat is not sleep.
the cat is sleep.
**any other cats? i can see only 1 cat.**
none
**any people? no.**
there are no people.
there are people.
**is it day? yes it is.**
none
**any sunshine? yes nice sunshine.**
there is a sunshine.
there is no sunshine.
**is this in a yard or park? it's a park.**
none
**is the field big? no in the picture.**
the field is not big.
the field is big.
**angry birds? i don't see any birds.**
none

Figure 10: Example of generated propositions for VisDial dialogues (CC-BY 4.0) from the training set, after downsampling captions and before balancing.

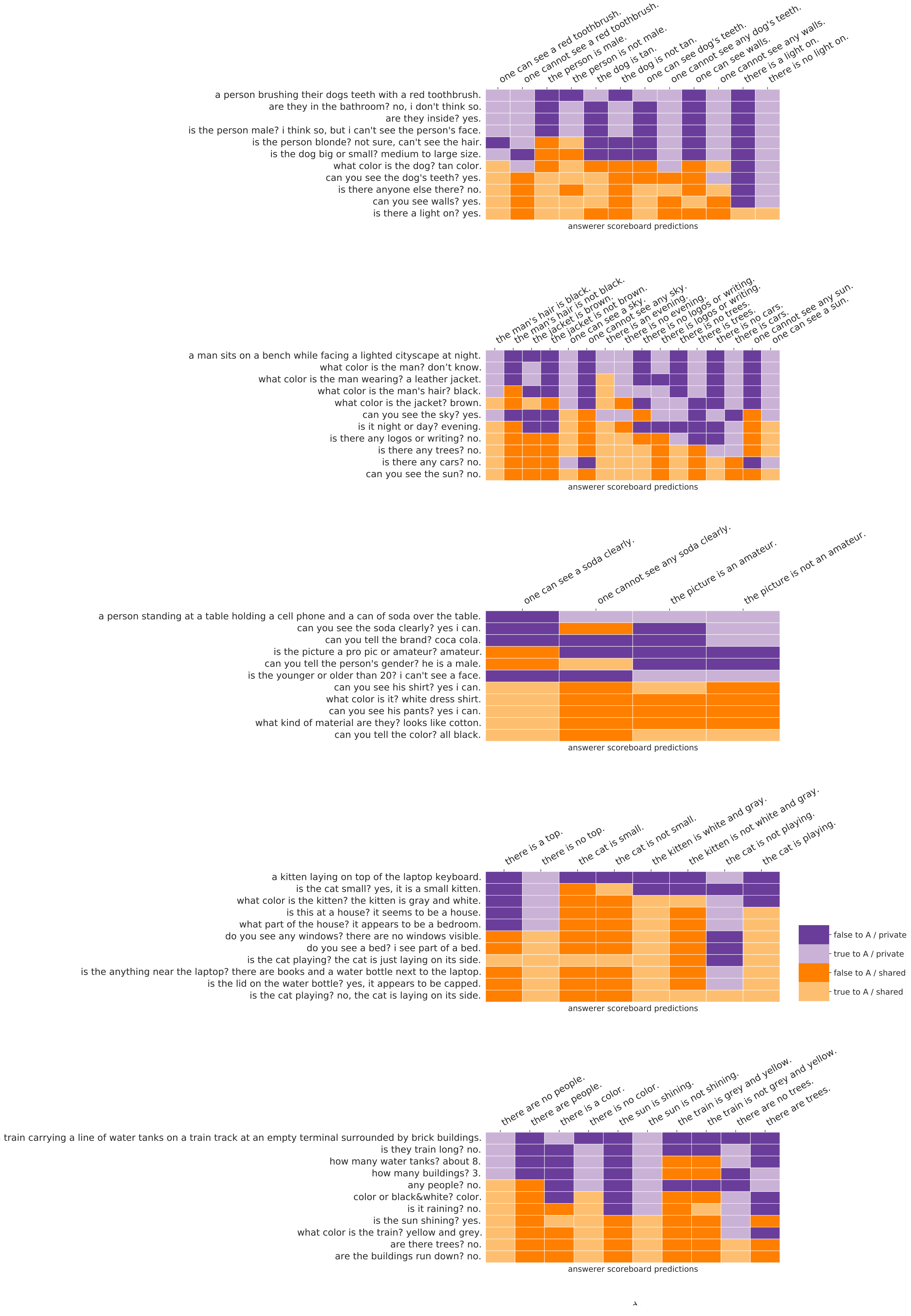

Figure 11: Examples of complete predicted scoreboards by $A$, main task, RL_DIV on TFxPS. All dialogues are from the VisDial validation set (CC-BY 4.0).